# Explaining Hate Speech Classification with Model-Agnostic Methods

Durgesh Nandini[1], Ute Schmid [2]

**Abstract:** There have been remarkable breakthroughs in Machine Learning (ML) and Artificial Intelligence (AI), notably in the areas of Natural Language Processing (NLP) and Deep Learning. Additionally, hate speech detection in dialogues has been gaining popularity among Natural Language Processing researchers with the increased use of social media. However, as evidenced by the recent trends, the need for the dimensions of explainability and interpretability in AI models has been deeply realised. Taking note of the factors above, the research goal of this paper is to bridge the gap between hate speech prediction and the explanations generated by the system to support its decision. This has been achieved by first predicting the classification of a text and then providing a post-hoc, model-agnostic and surrogate interpretability approach for explainability and to prevent model bias. The bidirectional transformer model BERT has been used for prediction because of its state-of-the-art efficiency over other Machine Learning(ML) models. The model-agnostic algorithm LIME generates explanations for the output of a trained classifier and predicts the features that influence the model's decision. The predictions generated from the model were evaluated manually, and after thorough evaluation, we observed that the model performs efficiently in predicting and explaining its prediction. Lastly, we suggest further directions for the expansion of the provided research work.

## 1 Introduction

In the past few years, there have been breakthroughs in machine learning, and artificial intelligence challenged the dimensions of explainability and interpretability. Explainable AI(XAI) will be essential if users are to understand, appropriately trust, and effectively manage the Artificially Intelligent systems [WS20].

Explainability and interpretability are often used with blurred edges of where interpretability ends and explainability begins. However, Molnar defines explainability as how well the model can explain its inner workings to human users [Ch20]. In contrast, interpretability is the association of the cause and effects of the model. Explainable and interpretable models are frameworks that inherently involve humans in the loop by providing the users with an understanding of the model behaviour and enabling the building of robust models with strengthened trust and improved decision-making capabilities due to transparency, usability, model and/or result in justification [Gi18; Gu18; LL17a; Mi19; TTN06].

[1] Otto-Friedrich Universität Bamberg, An der Weberei 5, 96047, Germany durgesh.nandini@uni-bamberg.de
[2] Otto-Friedrich Universität Bamberg, An der Weberei 5, 96047, Germany ute.schmid@uni-bamberg.de

Machine interpretability methods are often categorised along with three main criteria [AB18; Gi18]. The first discriminates based on the coverage of explanation as local or global for an explanation at instance-level (individual predictions) or model-level (entire model). Local interpretability methods explain how predictions change when input changes and are applicable for a single prediction or a group of predictions. Global interpretability methods simultaneously explain the entire ML model, from input to prediction. The second category of distinction is intrinsic vs post-hoc interpretability models. While intrinsic interpretable models are self-explanatory inherently and explanations are simply by-products of model training, post-hoc interpretability is realised only after external algorithms are applied to the trained models. Lastly, interpretability can also be categorised based on being model-agnostic or model specific. Model-specific means that a dedicated interpretability model is created for the training model. In contrast, model-agnostic interpretability methods can be plugged and customised with any Machine Learning model in use. XAI models may be deployed and used in various domains for real-time use. For instance, an XAI model in the domain of financial loans may explain why a certain user may be granted or denied a loan [KS20]. In this work, we analyse the domain of hate speech on social media.

Hate speech detection in dialogues has been gaining popularity among NLP researchers with the increased use of social media. What can be defined as hate speech is that it is understood to be bias-motivated, hostile and malicious language targeted at a person or group because of their actual or perceived innate characteristics [Br18; MSB17; SK22; Wa12; WS95]. Online hate speech is heterogeneous and dynamic: it takes many forms, which can shift and expand over a relatively short span of time. The characteristics that add to the dangers that hate speech poses are accessibility, diversity, instant reaction rates, anonymity and multiplicity [Br18]. This implies that while users are not compelled to reveal aspects of their offline identity unless they wish to do so [Br18; MSB17], the accessibility that they possess on the internet and social media platforms is readily high, with the size and reach of the audience is quite large, anybody and everybody can view, replicate, and support or replicate the hate speech, and because the access is so high, the reaction rate is also high be it in support of the hate speech or producing hate speech as a reply to the original hate speech [Br18]. The last characteristic, multiplicity, is related to the above point of reaction rate. Since there is no way to predict the reaction rate, there is no way to predict the magnanimous multiplicity the hate speech may result in [Br18].

The aforementioned forms the motivation for the work presented in this paper. This paper aims to predict and explain hate speech in tweets in the form of texts. The major goal of our work is to provide the basis for coherent, comprehensible, contextual, and realistic explanations with high local fidelity. This is done using the model-agnostic surrogate model approach. The model agnostic model Local Interpretable Model-agnostic Explanations (LIME) generates explanations for the output of a trained classifier and predicts the features that influence the decision of the model [RSG16]. Furthermore, the post-hoc interpretability approach has been used to prevent model bias [Ho21]. The interpretability methods show the features that influence the decision of the model the most and help the user understand how the model is deciding to detect hate speech in dialogues [RSS21]. The presented work provides insights into the decision points and feature importance used to make predictions about the hate speech disposition of conversations. The architecture provided in the paper is not limited to the text domain but can also be extended for multimodal settings.

The sections of the paper are organised as follows: Section 1 provides a walk-through of the taxonomy of the Explainable Artificial Intelligence (XAI), discussing the types of XAI models available and the paths that can be chosen when seeking different types of explainability, Section 2 discusses a few existing models for explainability and interpretability, and Section 3.1 then gives details about the data that has been used in this work, Section 3 provides a description of the methodology used in the paper and the architectural flow of the methodology used, Section 4 describes the experiments conducted for prediction of hate speech and the explanations provided for them, Section 5 describes the results obtained from the experiments and provides an evaluation of the model and the outcomes. Lastly, Section 6 has the conclusion and future works.

## 2 Related Work

This section provides a brief overview of canonical works and techniques relevant to our study. We first look at relevant areas of hate speech detection; then, we have a brief overview of the models that can be used to explain the results of the detected hate speech and the various modalities of the explainable and interpretable models.

A simple mode of hate speech detection is using a set of keywords and checking for the availability of the keywords in the text [Ma19; MSB17; SM22]. The use of word embeddings and a Bag-Of-Words (BoW) model is advancing a bit further. Kwok et al. [KW13] use a unigram model for hate speech detection and [Dj15] use embeddings for the same. [KW13] have also combined their BoW model with a Naive Bayes classifier. Another popular method of detection is combining NLP techniques with a ML model. MacAveney et al. [Ma19] in their method propose a multi-view SVM approach for hate speech detection. Furthermore, Deep Learning (DL) may also be used for the subject matter at hand, either with a sole method or in an ensemble of one or more methods. [Ba17; GS17; ZKF18] use Convolutional Neural Networks (CNN), [Ba17; De17] use Recurrent Neural Networks (RNN) and [MFC19; MFC20; SN20] use transformers for the same.

Having had an overview of the hate speech detection models, we now look at methods to introduce interpretability in our model. For this, we are going to go through SHAP [LL17a], anchors [RSG18] and LIME [RSG16].

Inspired by coalitional game theories, SHAP (SHapley Additive exPlanations) is an approach to explaining the prediction of an instance by computing the contribution of each feature to the prediction. The resulting SHAP values [LL17b] can be interpreted as a unified measure of additive feature importance [LL17a]. The primary advantages of SHAP are that it offers global interpretability by showing whether a variable has a positive or negative influence on the target value, as well as local interpretability by showing why a case receives its prediction and the contributions of the predictors. Although Shapley values fulfil desirable properties like local accuracy, missingness, and consistency, its model-agnostic approximation technique Kernel SHAP is slow concerning computational time complexity, on the one hand, [Ki22].

Ribeiro et al. [RSG18] propose a novel model-agnostic system that explains the behaviour of complex models with high-precision rules. The rules are called anchors and represent local, sufficient conditions for predictions. The authors claim that rule-based anchors provide an intuitive classification and prediction by highlighting the part of the input that suffices the prediction. An anchor explanation is a rule that sufficiently ties up the prediction locally – such that changes to the rest of the feature values of the instance do not matter. Because of the use of if-then rules, the explanations generated are easy to understand [RSG18]. However, the rule-based algorithm does not specify how should conflicting and contrasting rules be resolved or what should be the coverage area of the anchors.

Ribeiro et al. [RSG16] propose LIME, a model agnostic interpretability model that explains an individual model's prediction by locally approximating the model's decision boundary in the neighbourhood of the given instance. A major advantage of LIME is that each explanation is easy to understand, even for complex models and tasks. Still, the caveat is that the model can only capture the model's behaviour on a local region of the input space [RSG18]. The LIME methodology offers local fidelity, local interpretability via instance explanation, and global interpretability via model explanation. LIME uses a local linear explanation model and can thus be characterised as an additive feature attribution method.

State-of-the-art also gives an insight into the different levels and modes in which explanations might be generated and what may suggest an explanation to be complete. Kumar et al. [KDA21] suggest that explanations reveal a transactional nature and reflect an attempt to communicate an understanding between individuals. The authors also suggest that features play an exuberant role in predicting outcomes as well as generating explanations for interpretability. Another insight may be that explanations can be generated according to various levels of understanding. Bettina et al. [Fi21] describe three levels of explanations based on global, local or endless(theoretically) levels of details for a class of examples or models. What may be understood from this state-of-the-art literature is that explanations should stick together and represent an internally consistent package whose elements form an interconnected, mutually supporting relational structure [TTN06].

## 3 Methodology

Although black-box ML models have observable input-output relationships, they lack transparency into their internal functioning. This is typical of deep learning, which models very complex problems with high nonlinearity and input-input interactions [KDA21]. As such, it is essential to decompose the model into interpretable components to simplify the model's decision-making procedure.

In this section, we described the steps involved in generating explanations and their analysis. Our methodology builds on top of the classic LIME model and is characterised by the integration of a supervised machine learning model and an explanation system. Fig. 1 illustrates the architecture pertained in the paper. The following subsections discuss the steps applied in our methodology.

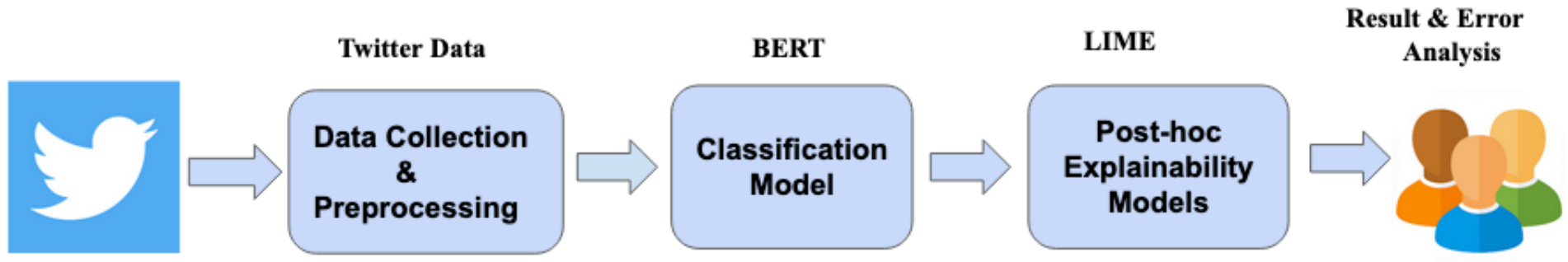

Fig. 1: A pipeline of the proposed Methodology

### 3.1 Data

We used a Twitter dataset for the case study in this work. The dataset was derived from Davidson et al. [Da17]. For the dataset, the organisers have collected a large archive of tweets using a Twitter API and pre-selected potentially problematic posts using lexicons from Hatebase.org[3]. The authors collected a total of 85.4 million tweets from 33,458 Twitter users and then they have taken a random sample of 25,000 tweets. The tweets are then judged and annotated by at least three humans. They evaluated the intercoder-agreement to be 92%. After the annotation, 24,802 labelled tweets were obtained as the authors suggest that some tweets were not assigned labels as there was no majority class. It is also interesting to note that the authors also indicate that the annotators were asked to label the data as per the context in which they were used, meaning that a mere presence of a word did not indicate the tweet to be a Hate or Offensive speech. The class distribution of data is shown in Fig. 2a. For our case study, we have used three classes from the dataset: 'Hate', 'Offensive' and 'None'. The three classes are defined below.

*Hate:* Describing negative attributes or deficiencies to groups of individuals because they are members of a group. There are hateful comments toward groups because of race, political opinion, sexual orientation, gender, social status, health condition, or similar. An example of this category would be äll poor people are stupid".

*Offensive:* Posts which are degrading, dehumanising, insulting an individual, or threatening with violent acts are categorised into this category. An example for this category would be "f**king forget that b***h".

[3] https://hatebase.org/

*None:* Posts that do not belong to any of the above categories are categorised in this set.

The data was highly imbalanced, the majority were in none class, and only 1,430 were hateful tweets, as shown in Fig. 2a, so we have used a random sample of 1,430 tweets from each class. Fig. 2b demonstrates a visual representation of the data as a word cloud. It can be seen that the most prominent words highlighted are a balanced mix of hate, offensive and none data. It can also be observed that a lot of hate words are being targeted toward people, communities, or organisations. This finding does conform to the aim of the research work focus. The classes mentioned are not mutually exclusive. A tweet may fall into one or more categories; however, the class of the tweet is dependent on the probability score of the different classes. The class with the highest probability score is assigned the tweet.

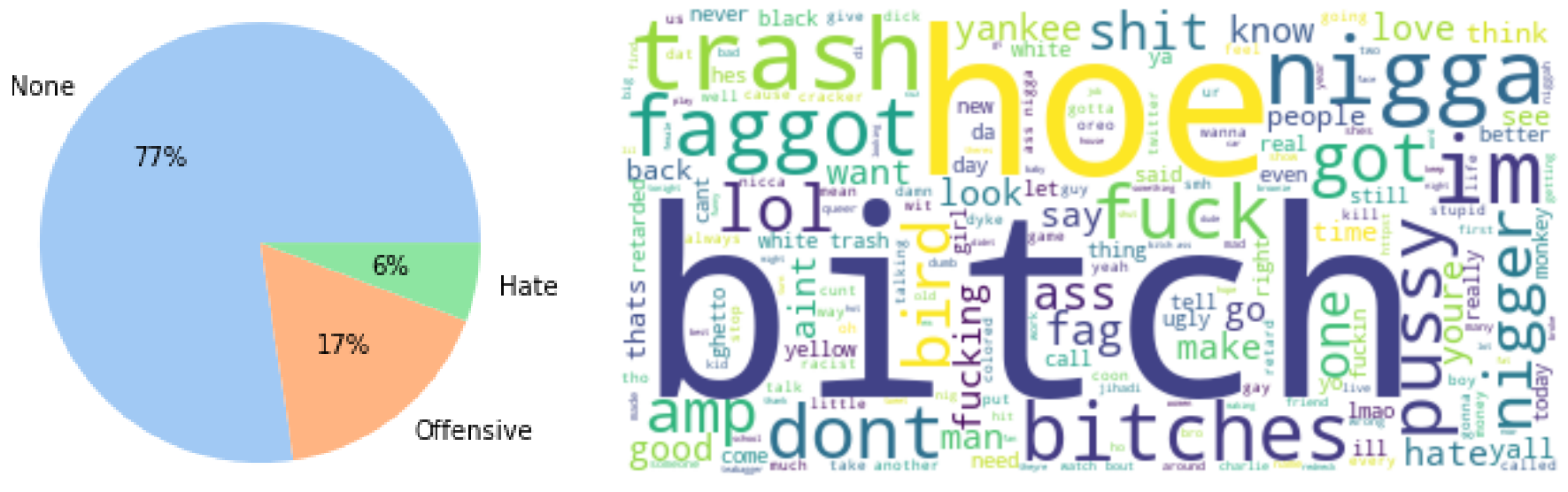

(a)

(b)

Fig. 2: a) Class distribution of all tweets b) Word Cloud of sampled tweets

### 3.1.1 Data Preprocessing

The language ongoing on social media is usually casually written with no special emphasis on grammar or literary correctness and in combination with emojis, symbols and hashtags. For our pipeline, we have preprocessed the data to remove smileys, emojis and any other symbol that may be present. In addition to that, we have also eliminated stopwords as the model performed almost the same when with or without the stopwords. The hashtags were not eliminated because we observed that the hashtags contributed to the meaning of sentences and would often encapsulate the emotions of sentences, hence contributing to categorising a sentence as hate speech or not.

## 3.2 Classification Model

The data is trained using the Bidirectional Encoder Representations from Transformers (BERT) model [De18]. BERT is a state-of-the-art NLP model that applies bidirectional training of attention mechanism to language modelling tasks. The bidirectional flow of training provides a deeper insight into the language context. In vanilla form, BERT is composed of an encoder that reads the text input, which may then be integrated with a classification model to predict a task. Unlike directional models, which read the text input sequentially (left-to-right or right-to-left), the Transformer encoder in BERT reads the entire sequence of words simultaneously. This characteristic allows the model to learn the context of a word based on all of its surroundings.

### 3.3 Post-hoc Explainability Models

Post-hoc interpretability approaches propose to generate explanations for the output of a trained classifier in a step distinct from the prediction step. These approximate the behaviour of a black box by extracting relationships between feature values and the predictions. Two widely accepted categories of post-hoc approaches are surrogate models and counterfactual explanations. Surrogate model approaches aim to fit a surrogate model to imitate the behaviour of the classifier while facilitating the extraction of explanations. Often, the surrogate model is a simpler version of the original classifier. Global surrogates are aimed at replicating the behaviour of the classifier in its entirety. On the other hand, local surrogate models are trained to focus on a specific part of the rationale of the trained classifier. This research uses the post-hoc local surrogate explainability method, Local Interpretable Model-agnostic Explanations (LIME) [RSG16]. Typically, LIME creates a new dataset or text from the original text by randomly removing words from the original test and gives the probability to each word to eventually predict based on the calculated probability.

#### 3.3.1 *LIME*

LIME is a local surrogate approach that approximates any black-box machine learning model with a local, interpretable model to explain individual prediction. The model specifies the importance of each feature to an individual prediction. The model works by tweaking the inputs slightly and observing the changes in prediction. The tweaked data points are weighed as a function of their proximity to the original data points, then fitting a surrogate model such as linear regression on the dataset with variations using those sample weights. Each original data point can then be explained with the newly trained explanation model [RSS21]. The learned model generates a local prediction model while it may or may not provide a precise global approximation. Since LIME models treat the machine learning models as a black box, these are model agnostic. The Fig. 3 shows a flowchart of the LIME methodology.

Mathematically, LIME explanations are determined using the formula:

$$\xi(x) = argminL(f, g, \Pi_x) + \Omega_g; g\epsilon G, \tag{1}$$

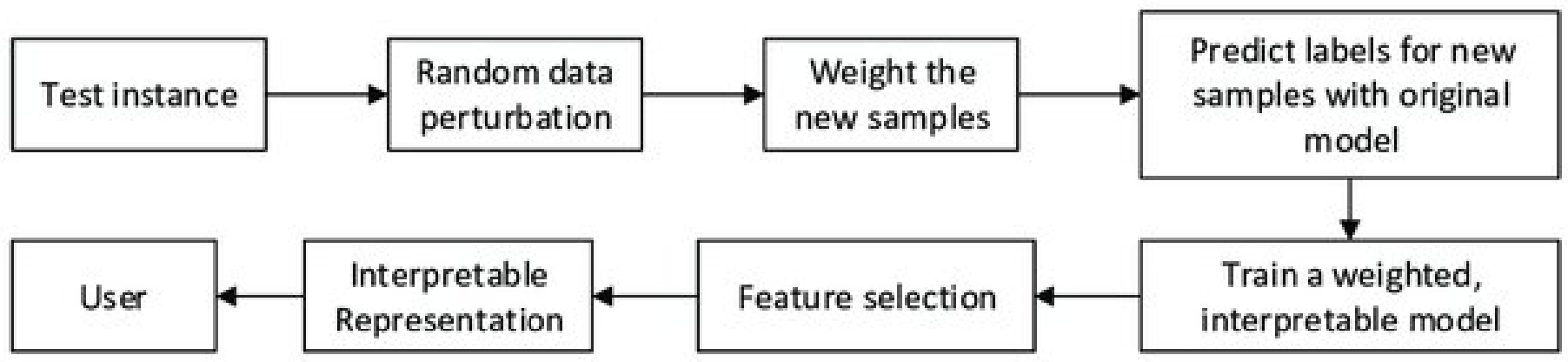

Fig. 3: Flowchart of the LIME methodology[ZK21]

The mathematical formula above states that the explanation for a data point is the model that minimises the locality-aware loss keeping the model complexity low. The loss function L measures the closeness of the explanation to the prediction of the original model f while keeping the model complexity $\Omega_g$ low. G is the pool of possible explanations, and $\Pi_x$ is the proximity measure of how large the neighbourhood is around the instance x. LIME optimises only the loss part of the data [RSG16].

The idea for training the LIME model is presented in Fig. 3 :

- Select the instance in which the user wants to have an explanation of the black-box prediction
- Add a small noisy shift to the dataset and get the black-box prediction of these new points
- Weight the new point samples according to the proximity of the instance x
- Weighted, interpretable models are trained on the dataset with the variations
- With the interpretable local model, the prediction is explained

### 3.4 Error Analysis

ML models often face evaluation challenges regarding performance, accuracy, and reliability. In practice, there might be a possibility that the model accuracy may not be uniform across subgroups of data and that input conditions might exist for which the model fails. Henceforth, we analyse the results obtained from the experiments and draw meaningful conclusions from the results obtained. To achieve this, we perform an error analysis to evaluate the performance of the classification and the explanation model.

## 4 Experimental Setup

For experimental purposes, we have used the tweets collected from Twitter. The data is then preprocessed following the preprocessing pipeline as mentioned in Section 3.1.1. The stopwords and pronouns were removed, followed by lemmatisation and stemming. We have used the NLTK library[4] for the removal of stopwords and customised it to remove additional pronouns from our corpus. The lemmatisation and stemming were performed using the NLTK WordNet lemmatiserand stemmer. The hashtags of tweets are preserved as they contribute meaningfully to our experimental scenario. The preprocessed data is then split into a training set, a test set and a validation set in a standard ratio of 70:20:10. The data is then labelled into their respective classes for training purposes. The labels are Hate, Offensive and None. Details about the data are provided in Section 3.1. Supervised learning is carried on to the preprocessed and labelled training data for the next step. For the supervised learning classification, we use a BERT model.

After supervised learning, we then add the LIME pipeline to the prediction outcomes from the supervised learning model. We perform all our experiments using the Python language.

## 5 Results and Discussion

This section observes and evaluates the results obtained from the experiments conducted. After preprocessing the data, we use the BERT classification model for supervised learning. A detailed classification report is given in Tab. 1. We run the classification model for four epochs as the performance stabilises after the four epochs. The result presents the precision, recall, F1 and accuracy scores rounded to the third decimal digit. The best accuracy obtained is 82.6%. Then we add the LIME architecture to the pipeline and evaluate the LIME results. The bottom right part of Fig. 4 shows an example tweet from our dataset.

Tab. 1: Summary of the classification report

| **Epoch** | **Precision** | **Recall** | **Accuracy** | **F1 Score** |
|---|---|---|---|---|
| 1 | 0.819 | 0.824 | 0.819 | 0.820 |
| 2 | 0.818 | 0.817 | 0.815 | 0.817 |
| 3 | 0.824 | 0.826 | 0.823 | 0.826 |
| 4 | 0.832 | 0.814 | 0.826 | 0.828 |

In our experiments, LIME measures similarity between documents using similarity kernels that are based on distance measures. Context and coherent semantic facts that should reveal a similar meaning are analysed as basic explanation units regarding their effect on the classification when being removed. The most relevant classes with their assigned words are presented as explanations. Currently, a manual evaluation technique has been employed for evaluation purposes where a human evaluator checks the decision made by the model and the explanation generated by LIME and assesses their accuracy. In the following paragraphs, we analyse and discuss the experimental results.

[4] https://www.nltk.org/index.html

As discussed in Section 3, we have used supervised learning to generate a one vs all classification and then interpreted the results via LIME. The classification algorithm generates a probability distribution for each class and a probability score for each feature per class. This essentially means that there was an overlap of features for one or more classes, but the model decision was made in favour of the class that had a maximum hold on the feature scores. We show an example result to shed light on this discussion.

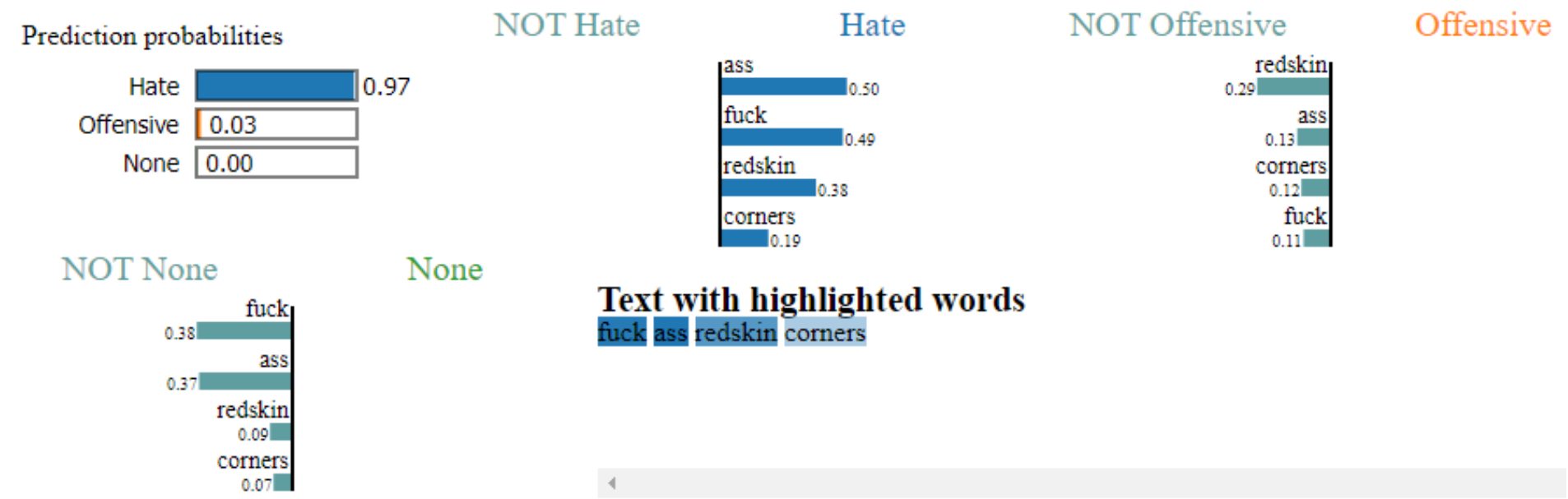

Fig. 4: Hate Prediction by the BERT classification model along with explanations generated by the LIME pipeline

Fig. 4 represents a one vs all predictions generated by the LIME classification algorithm and depict the features that were deemed relevant for the decision-making by the algorithm. Fig. 4 also illustrates the feature probability distribution for each class. The Prediction Probability in Fig. 4 shows the per class probability distribution, and the Text with highlighted words illustrates the features that were used to make the decision. As displayed, the feature äss"has a score of 0.5, "f**k"has a score of 0.49 and "redskin"has a score of 0.35; depending on which the classification decision for the 'hate' class was decided upon by the model. These features explain the decision of the classifier.

To evaluate errors in the model, we start by observing 150 tweets, taking 50 random tweets from each class. Overall, we identified a 21% error rate of our 150 tweet texts, where 5% were predicted as false positives, and 16% were predicted as false negatives, where false negatives in our case would be an incorrect classification of tweets. In the observed number of tweets, a mere 0.6% tweets had an exactly equal probability of 'hate' and 'offensive'. For the Hate class, 78% of the analysed tweets were correctly classified, whereas 18% were incorrectly classified as Offensive, 4% were incorrectly classified as None class. Similarly, for the Offensive category, 74% of the analysed tweets were correctly classified, while 22% were incorrectly classified as Hate, 2% were incorrectly classified as None, and 2% shared an equal probability of Offensive and Hate. For the None class, 84% of the analysed tweets were correctly classified, whereas 2% were incorrectly classified as Offensive, and 14% were incorrectly classified as Hate. In diagnosing the predicted tweet texts and their classes, we identified a few words that always caused the results to fall into a particular category.

We also observed that certain words had a higher frequency of occurrence in each class. The top sixty frequent words for the class Hate, Offensive and None are shown in Fig. 5a, 5b and 5c respectively.

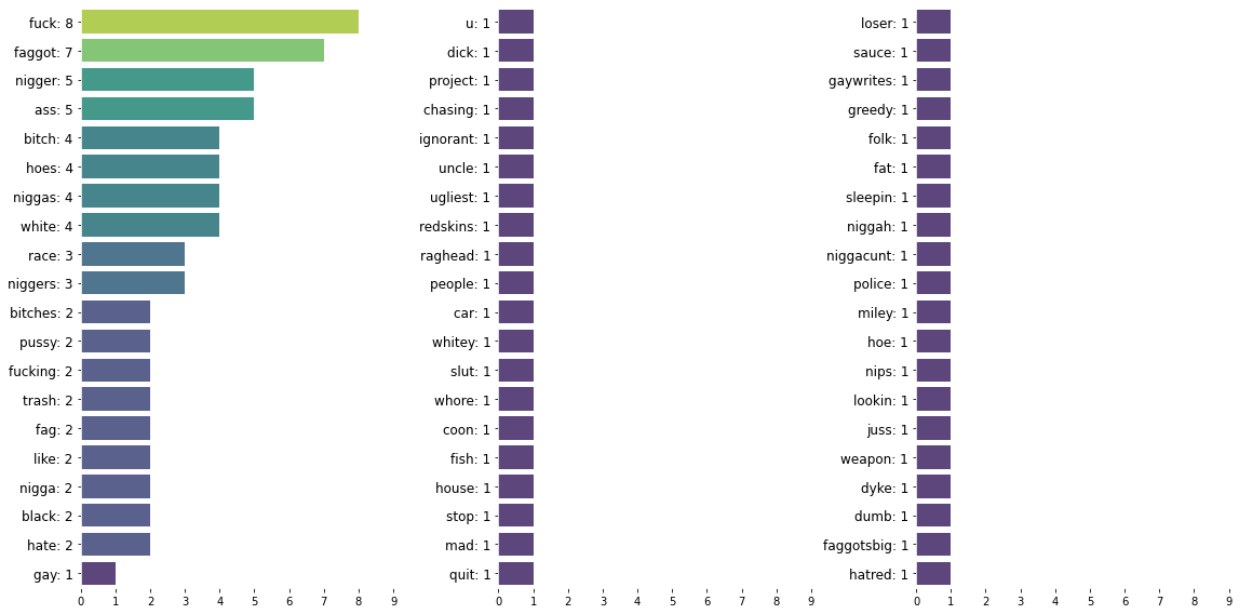

(a)

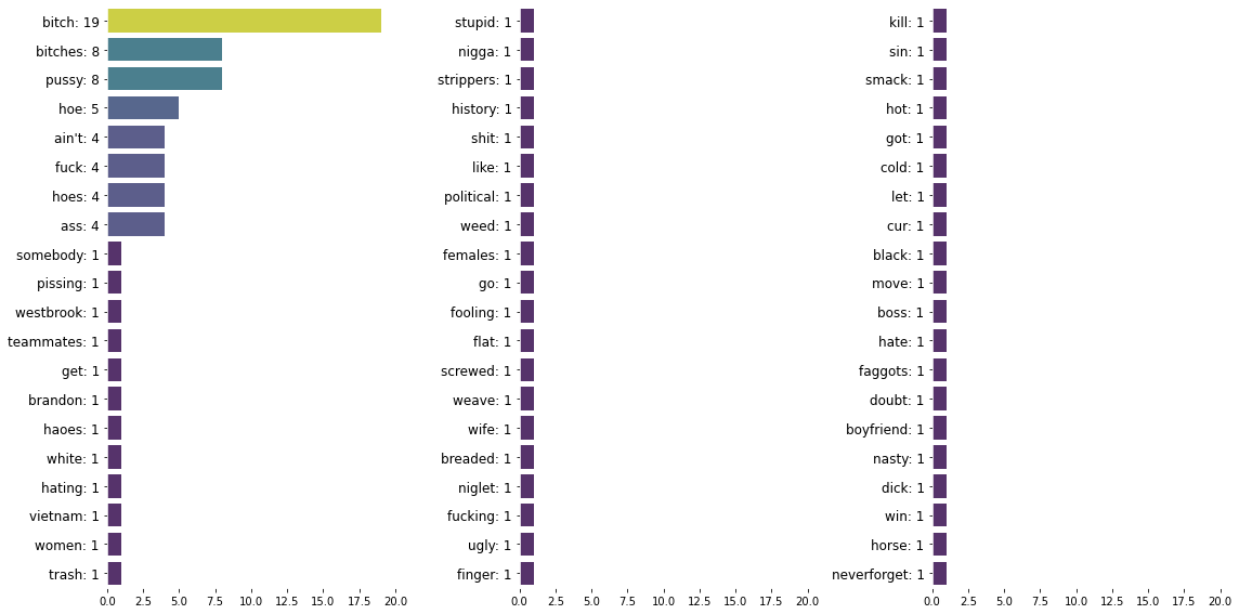

(b)

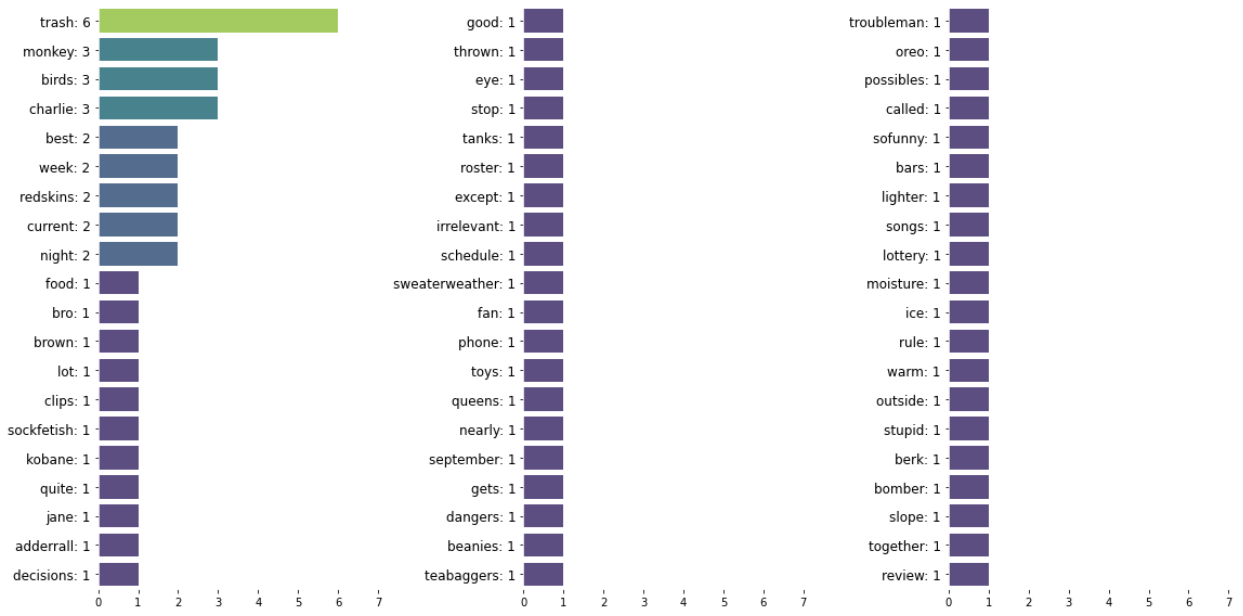

(c)

Fig. 5: Frequent words obtained from different classes a) Hate b) Offensive c) None

While evaluating the results, we observed that while there may be a feature overlap among classes, the model decision was the effect of a combined factor of hate words, nouns, and pronouns. For example, if the features were directly using the words like 'you are  hate word (from the shown hate word list, for example)', then the model decision was in favour of the class Hate. At the same time, the tweet text was classified as offensive if there were an indirect reference or no direct pointing of objects. Additionally, if the tweet used an offensive word but discussed abstract universal concepts, then the tweet was classified as 'offensive as well. If the tweet had no offensive or hate words, then the model classified them as 'none'. For instance, Fig. 5 show that the words like "f**k", "b***h", "black", are frequently occurring in Hate, None, as well as Offensive classes. Here the model checks other combinatorial words and then decides whether a tweet is Hate or Offensive. So, if the word directly refers to a Noun, Pronoun, for example, Ï'd f**k a dog before I f**k you fish black p***y", then the tweet was classified as Hate, while äye yo black car is superior"was classified as None. Another conclusion from the analysis was that tweets were almost always categorised as Hate when they were racist (use of stem words such as "black", "n***a", "f****t", "white", etc.), while they were almost always classified as Offensive when they were sexist (use of stem words such as "cunt", "b***h", "hoe", "p*****g wife", "p***y", etc.).

## 6 Conclusion and Future Work

In this work, we shed light on the rising effects of hate speech on social media and the dangers that they pose due to various factors. To solve this issue, we propose a methodology to detect hate speech on social media platforms and provide an explanation for the same using feature vectors. We have worked on the Twitter dataset for experimental purposes.

This work opens the prospects for numerous future works, such as enriching the architecture with rule-based learnings using named entity recognition (NER) in association with relational features. The architecture can also be extended to various dimensions of data, for example, using image data, spatial relations in textual or image data, or both. Moreover, to evaluate how human users evaluate the model, a survey can be conducted to evaluate the prediction outcomes or the explanations.

## 7 Acknowledgements

The research presented here has been carried out within the Dare2Del project under the DFG priority program Intentional Forgetting (SPP 1921). Dare2Del is a joint project of Cognitive Systems, University of Bamberg and the Chair for Work and Organisational Psychology, University of Erlangen.